\pdfoutput=1

\documentclass[11pt]{article}

\usepackage[final]{coling}

\usepackage{times}
\usepackage{latexsym}
\usepackage{microtype}
\usepackage{hyperref}
\usepackage{url}
\usepackage{booktabs}
\usepackage{graphicx}
\usepackage{amsmath}
\usepackage{multirow}
\usepackage{makecell}
\usepackage{subcaption}
\usepackage{listings}
\usepackage{courier}
\usepackage{wrapfig,lipsum,booktabs}

\usepackage{tcolorbox}
\definecolor{bb}{HTML}{95e1d3}
\definecolor{gg}{HTML}{c7ffd8}
\definecolor{yy}{HTML}{f0c38e}
\definecolor{blu}{HTML}{5ab4ba}
\definecolor{rr}{HTML}{f38181}

\definecolor{c1}{HTML}{6E85B2}
\definecolor{c2}{HTML}{368B85}
\definecolor{c3}{HTML}{C56824}
\definecolor{c4}{HTML}{FFC069}
\definecolor{c5}{HTML}{916BBF}

\usepackage[T1]{fontenc}

\usepackage[utf8]{inputenc}

\usepackage{microtype}

\usepackage{inconsolata}

\usepackage{graphicx}

\title{
Large Language Models for Predictive Analysis: How Far Are They?
}

\author{
 \textbf{Qin Chen\thanks{Equal contribution.} \textsuperscript{,1}},
 \textbf{Yuanyi Ren\footnotemark[1] \textsuperscript{,1}},
 \textbf{Xiaojun Ma\thanks{Corresponding author.}\textsuperscript{,1}},
 \textbf{Yuyang Shi\textsuperscript{2}},
\\
 \textsuperscript{1}Peking University,
 \textsuperscript{2}Harvard University,
\\
\small \texttt{
\{chenqink, yyren, mxj\}@pku.edu.cn , yuyangshi@fas.harvard.edu
}
}

\begin{document}
\maketitle
\begin{abstract}
Predictive analysis is a cornerstone of modern decision-making, with applications in various domains. Large Language Models (LLMs) have emerged as powerful tools in enabling nuanced, knowledge-intensive conversations, thus aiding in complex decision-making tasks. With the burgeoning expectation to harness LLMs for predictive analysis, there is an urgent need to systematically assess their capability in this domain. However, there is a lack of relevant evaluations in existing studies. To bridge this gap, we introduce the \textbf{PredictiQ} benchmark, which integrates 1130 sophisticated predictive analysis queries originating from 44 real-world datasets of 8 diverse fields. We design an evaluation protocol considering text analysis, code generation, and their alignment. Twelve renowned LLMs are evaluated, offering insights into their practical use in predictive analysis. Generally, we believe that existing LLMs still face considerable challenges in conducting predictive analysis. See  \href{https://github.com/Cqkkkkkk/PredictiQ}{Github}.

\end{abstract}

\section{Introduction}
\label{sec:intro}
Predictive analysis \citep{predictiveAnalysis} involves making predictions about future outcomes based on past data, using statistical modeling, data mining, and machine learning techniques. It is widely used in decision-making across various fields like business \citep{che2024integratinggenerativeaifinancial} and healthcare \citep{ Dixon2024UnveilingTI}.

\begin{figure*}[h]
    \centering
    \includegraphics[width=\linewidth]{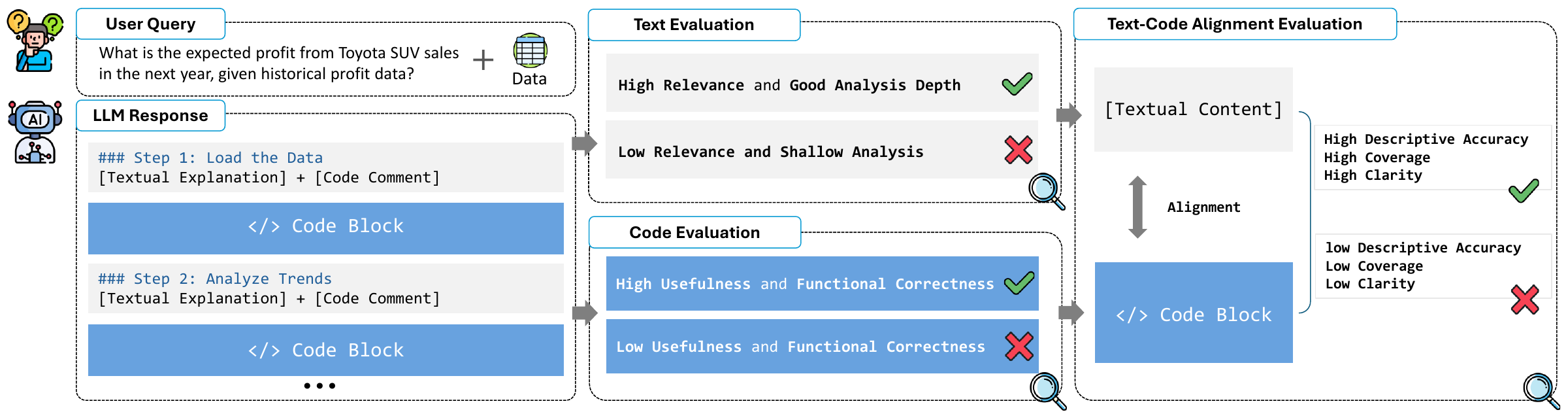}  
    \caption{An example of users conducting predictive analysis via Large Language Models.}
    \label{fig:user-llm-example}
\end{figure*}

To facilitate off-the-shelf predictive analysis for users without data analysis expertise, Large Language Models (LLMs) \citep{brown2020language,openai2023gpt4,anil2023palm,abdin2024phi} serve as powerful tools, supporting users with coherent and contextually relevant responses. 
\autoref{fig:user-llm-example} shows how users engage with LLMs by submitting predictive queries based on specific datasets.

Given the substantial potential of LLM-based predictive analysis, a comprehensive evaluation of leading LLMs is essential. While no studies specifically assess LLM-based predictive analysis, related studies \cite{zhao-etal-2023-investigating,chen2022large,saeed2023querying,gao2023texttosql,he2023text2analysis,ABOLGHASEMI2024,pratt2024languagemodelsuseforecasting, hong2024datainterpreterllmagent} in relevant fields primarily focus on evaluating either the model's outcomes only (e.g., a total sales figure answering a database query) or the generated codes that, when executed, produce these outcomes. 
However, the former raises scalability issues as LLMs may struggle to process the entire lengthy dataset due to limited context length. The latter lacks textual explanations, such as justifications for algorithm selection, which constrains its applicability and undermines user trust. This makes existing evaluation efforts insufficient when directly adapted to predictive analysis, where tasks are inherently more complex.

LLM-based predictive analysis typically requires handling tasks such as data preprocessing, algorithm selection, result interpretation, and so on.
These tasks require both \textbf{textual explanations} and \textbf{code implementations}. In practical applications, textual analysis is crucial for explaining algorithm selection and enhancing reliability. Moreover, effective alignment between text and code improves user comprehension of LLM-generated solutions. Thus, \textbf{textual analysis}, \textbf{code generation}, and \textbf{text-code alignment} are integral to the functionality of AI-driven assistants.

In this paper, we propose the PredictiQ (Q stands for query) benchmark to evaluate the potential of current LLMs in predictive analysis. We first collect datasets covering eight common fields, including economics, traffic, and more. Diverse datasets form the basis for PredictiQ to assess LLMs' performance in predictive analysis. Based on the collected datasets, we collaborate with data analysis experts to formulate predictive queries tailored to the datasets. Each query is limited to a single, self-contained question that relies solely on the dataset, without requiring external knowledge. We then formulate the data-specific queries and corresponding datasets into prompts, which are provided to LLMs to generate responses. We propose an evaluation protocol encompassing three domains: textual analysis, code generation, and text-code alignment, focusing on seven key aspects. The responses from the LLMs are then evaluated by (\romannumeral 1) data analysis experts and (\romannumeral 2) LLMs. The LLM whose responses align most closely with those of the human experts is selected as the primary evaluator. In this study, we adopt GPT4Turbo, which achieves an average alignment ratio of 90.5\% with human experts. PredictiQ involves 1130 queries from 44 datasets across 8 fields, requiring 300 human hours for query generation and examination. Our evaluation on PredictiQ costs 900 human hours for response evaluation, 72.18 million input tokens, and 20.4 million output tokens.

We conduct extensive experiments on PredictiQ with twelve renowned LLMs. We find that (\romannumeral 1) Fine-tuning LLMs on code enhances both text analysis and code generation, occasionally pushing the model beyond its parameter limits. (\romannumeral 2) Code generation and text analysis are interconnected processes that shape LLMs' overall predictive analysis proficiency. 
(\romannumeral 3) Several LLMs' predictive analysis abilities vary significantly across different fields, with wide margins, and exhibit diversified score distributions.
We believe existing LLMs are far from mastering predictive analysis, both in terms of performance and efficiency.
Our contributions are summarized as follows.
\begin{itemize}
\item We propose the PredictiQ benchmark—a comprehensive evaluation framework for LLMs in predictive analysis. It covers the entire analytic process by integrating 44 datasets across 8 real-world fields, 1,130 rigorously curated data-specific queries, and evaluation protocols. This approach goes beyond existing benchmarks that focus only on basic descriptive tasks. 
\item We systematically evaluate the performance of twelve LLMs on the PredictiQ benchmark, highlighting insights that could  improve the efficacy of LLMs in predictive analysis.
\end{itemize}

\section{Preliminary}
\label{sec:Problem Definition}

\paragraph{Predictive Analysis} refers to the use of statistical algorithms and machine learning techniques to analyze data patterns, predict future events, trends, or behaviors \cite{siegel2013predictive,kumar2018predictive}. 
It is widely used in various fields, including finance, marketing, healthcare, and risk management, to anticipate potential risks, identify opportunities, and inform the decision-making process.
See \autoref{fig:user-llm-example} for an illustration.

\paragraph{Problem Definition.}

The predictive analysis process with LLMs is formalized as follows: 
\begin{equation}
    (\text{query}, \text{data}) \xrightarrow{\text{LLM}} (\text{text}, \text{code}).
\end{equation}
The input consists of a predictive analysis \textit{query} associated with corresponding \textit{data}. 
A \textit{query} is a data-specific predictive question that requests future predictions, identifies implicit data patterns, and so on.
The \textit{data} consists of multiple columns and rows.
The output includes \textit{text}, which explains how the prediction is made and justifies the chosen method, and \textit{code}, which implements the analysis. Ideally, the \textit{text} should provide clear and in-depth analysis based on the \textit{query} and \textit{data}. The \textit{code} should provide related and functionally correct implementation that solves the problem raised by the \textit{query}.
\section{The PredictiQ Benchmark} 
This section outlines the details of the PredictiQ benchmark, covering (i) data collection, (ii) query formulation, and (iii) response evaluation.

\subsection{Data Collection}

The PredictiQ benchmark evaluates LLMs' predictive analysis capabilities across diverse fields. It includes various publicly available datasets from real-world scenarios, ensuring the benchmark's diversity and validity. The datasets are organized into eight fields that are prevalent in real-world predictive analysis applications, as presented in \autoref{table:data-query-statistics}. See \autoref{sec:appendix_datasets} for detailed descriptions. 

\subsection{Query Collection}

\begin{table}[h]
    \centering
    \caption{Statistics of datasets and generated queries. }
    \label{table:data-query-statistics}
        \begin{tabular}{lccccccc}
            \toprule
            Field & \#Datasets& \#Queries \\
            \midrule
            Economics & 12 & 270 \\
            Marketing and Sales & 6 & 200 \\
            Industry Analysis & 7 & 180 \\
            Traffic & 5 & 130 \\
            Healthcare & 4 & 130 \\
            Social Study & 4 & 110 \\
            Human Resource & 3 & 80\\
            Education & 3 & 70\\
            \midrule
            \textbf{Total} & 44 & 1130 \\
            \bottomrule
        \end{tabular}
\end{table}

In this subsection, our goal is to formulate predictive questions tailored to each dataset. To achieve this, we engaged data science experts to develop well-defined queries following our instructions and examples. 
The provided instructions clarify the context of predictive analysis and outline the constraints on expected outcomes, as detailed below.

\textbf{Context of Predictive Analysis.} Based on our research into the application of predictive analysis across various fields \citep{business,marketing,healthcare,liu2023treeman}, we categorize common questioning methods into two types: those focused on predicting future trends and those focused on identifying previously unrecognized patterns within the dataset. We also encourage participants to consider queries related to time-series data.

\textbf{Constraints on Expected Outputs.} To ensure the quality of the outputs, we have implemented three specific instructions:
(\romannumeral 1) Participants are required to articulate a clear predictive target using unambiguous language. 
(\romannumeral 2) Queries should be based solely on the information within the dataset, excluding any reference to external data that could compromise the evaluation’s integrity. 
(\romannumeral 3) Each query should be restricted to a single question, ensuring a more balanced distribution of workload across all queries.
In addition, to improve real-world relevance and adaptability, participants are provided with background information on the sources and construction methods of the dataset.


To ensure diversity, we engage experts to select unique questions that cover distinct aspects of the dataset. The number of queries for each dataset is then tailored to its informativeness.
For further details, see \autoref{sec:appendix_datasets}. In total, we have generated 1130 queries from 44 datasets, covering common application scenarios in predictive analysis. The process of generating and examining the queries requires roughly 300 human hours.

\subsection{Response Generation} 

With the datasets and collected queries, we organize the data to formulate the input prompts. Each prompt consists of three parts: 
\begin{itemize}
    \item \textbf{Query and Instruction:} We instruct the LLM to perform predictive analysis, assuming the role of a professional data scientist.
    
    \item \textbf{Data Summary:} We provide a list of all columns in the dataset with their respective data types (e.g., \texttt{int}). We also include details such as the maximum and minimum values for numerical columns and the total number of categories for categorical columns. These summaries help the LLM better understand the dataset.
    
    \item \textbf{Data Details:} The dataset is provided in a CSV format, with columns separated by commas and rows by line breaks. This detailed format gives the LLM the essential information needed for predictive analysis.
\end{itemize}

The prompts are then submitted to the LLMs to generate responses. An example of a prompt and its corresponding response is provided in \autoref{sec:case_study}.

\subsection{Evaluation Protocols}

\label{sec:eval_protocol}

As mentioned in \autoref{sec:intro}, LLM-based predictive analysis involves both code and textual descriptions. It is essential to evaluate all three components: code generation, textual explanations, and the alignment between the two. Textual analysis offers key insights that enhance reliability, such as justifying algorithm choices, while code implementation operationalizes the analysis. Proper alignment helps users better understand the solutions generated by LLMs.
Building on this, we propose an evaluation protocol spanning three domains and seven aspects, each rated on a scale from 0 to 4. A summary is provided below, with full details available in \autoref{table:evaluation}.

\textbf{Text Analysis.} The text generated by LLMs is evaluated on two aspects: \textit{Relevance}, which assesses how closely the analysis aligns with the given data and the specific question, and \textit{Depth}, which evaluates the comprehensiveness of the justification for selecting a particular model or algorithm. We expect the textual analysis to provide a detailed and thorough examination of the analysis process, tailored to the specific data and queries.

\textbf{Code Generation.} This domain focuses on the quality of the code produced by LLMs with two aspects. Firstly, \textit{Usefulness} evaluates how well the code snippet addresses the given problem. Secondly, \textit{Functional Correctness} examines the code's execution correctness. We expect the generated code to accurately and effectively implement the associated predictive analysis.

\textbf{Text-Code Alignment.} We further evaluate the congruence between the generated code and textual analysis across three aspects: (\romannumeral 1) \textit{Descriptive Accuracy}, which assesses how precisely the text reflects the code; (\romannumeral 2) \textit{Coverage}, evaluating whether the generated text addresses all relevant aspects of the code, including its functions and nuances; and (\romannumeral 3) \textit{Clarity}, which examines the clarity of the alignment between the code and text. We expect the generated text and code to be well-aligned, and comprehensive, enhancing the user's understanding of the entire analysis process.

The proposed protocol offers a comprehensive evaluation of the responses. Advanced LLMs enable large-scale, stable, and effective assessment of the responses. To investigate how different LLM evaluators agree with experts' ratings and choose the appropriate evaluator, we employ (\romannumeral 1) human experts and (\romannumeral 2) LLMs to assess the responses according to this protocol. We then compare the score distributions from both groups to determine which LLM aligns most closely with human experts. Among the LLMs tested — GPT4Turbo, GPT4O, and Phi3Medium — GPT4Turbo demonstrated the highest alignment with human evaluations, as shown in our experiment in \autoref{fig:human-eval}. Consequently, we adopt GPT4Turbo as the primary evaluator. For assessing \textit{Functional Correctness}, we manually execute the generated code and evaluate the ratio of successful executions without errors.
\section{Evaluation Results on PredictiQ}

\subsection{Involved LLMs}
We evaluate eight popular LLMs on PredictiQ. 
(\romannumeral 1) \textbf{GPT Family.} GPT3.5Turbo, GPT4Turbo, GPT4O, GPT4O1, and GPT4O3Mini \citep{brown2020language,openai2023gpt4,openai2024gpt4ocard} are potent large-scale language models fine-tuned for both chat and code generation. 
(\romannumeral 2) \textbf{Llama Family.} This family includes CodeLlama2-7B, ChatLlama2-7B, ChatLlama2-13B, and ChatLlama2-70B \citep{touvron2023llama}. CodeLlama2-7B is fine-tuned for both chat and code generation. Other versions are reported to be fine-tuned for chat. 
(\romannumeral 3) \textbf{Other Models}. We also incorporate other popular LLMs, including Phi3Medium \citep{abdin2024phi3technicalreporthighly}, Phi4 \citep{abdin2024phi4technicalreport} and CohereRPlus\footnote{https://docs.cohere.com/v2/docs/command-r-plus}, into our evaluation. For LLM settings, please refer to \autoref{sec:settings}.

\subsection{Evaluation on PredictiQ}
\label{sec:eval_results}
\begin{table*}[htbp]
    \centering
    \caption{Evaluation results on PredictiQ across seven perspectives (scored 0–4) outlined in \autoref{table:evaluation}. \textit{Functional Correctness} represents the ratio of executable code, mapped to a 0-4 scoring scale. We also report average token costs, with additional reasoning costs for GPT4O1 and GPT4O3Mini.}
    \label{table:main-evaluation}
    \resizebox{.999\linewidth}{!}{
    \begin{tabular}{lccccccccc}
        \toprule
        & \multicolumn{2}{c}{\textbf{Text}} & \multicolumn{2}{c}{\textbf{Code}} & \multicolumn{3}{c}{\textbf{Text-Code Alignment}} & \multirow{3}{*}{\textbf{Total}} & \multirow{3}{*}{\textbf{Token Cost}} \\
        \cmidrule(lr){2-3}\cmidrule(lr){4-5}\cmidrule(lr){6-8}
         \multirow{2}{*}{LLM} & \multirow{2}{*}{Relevance} &  \multirow{2}{*}{Depth} &  \multirow{2}{*}{Usefulness} & Functional & Descriptive &  \multirow{2}{*}{Coverage} & \multirow{2}{*}{Clarity} &  \\ 
         & &  &  & Correctness & Accuracy &  & \\
        \midrule
        GPT3.5Turbo & 3.00{\tiny$\pm0.96$} & 1.76{\tiny$\pm0.69$} & 2.40{\tiny$\pm0.86$} & 2.12 (53\%) & 2.66{\tiny$\pm1.01$} & 2.47{\tiny$\pm0.94$} & 2.80{\tiny$\pm0.96$} & 17.21{\tiny$\pm5.19$} & 1934.34 \\
        GPT4Turbo & 3.39{\tiny$\pm0.79$} & 2.18{\tiny$\pm0.64$} & 2.78{\tiny$\pm0.72$} & 3.12 (78\%) & 3.09{\tiny$\pm0.80$} & 2.95{\tiny$\pm0.77$} & 3.18{\tiny$\pm0.77$} & 20.68{\tiny$\pm4.23$} & 2072.95 \\
        GPT4O & 3.60{\tiny$\pm0.65$} & 2.39{\tiny$\pm0.71$} & 3.12{\tiny$\pm0.67$} & 3.24 (81\%) & 3.36{\tiny$\pm0.68$} & 3.31{\tiny$\pm0.67$} & 3.41{\tiny$\pm0.64$} & 22.43{\tiny$\pm3.82$} & 3390.74 \\
        GPT4O1 & 3.61{\tiny$\pm0.51$} & 2.80{\tiny$\pm0.53$} & 3.45{\tiny$\pm0.60$} & 3.40 (85\%) & 3.47{\tiny$\pm0.68$} & 3.48{\tiny$\pm0.67$} & 3.48{\tiny$\pm0.67$} & 23.70{\tiny$\pm3.40$} & 6534.45 \\
        GPT4O3Mini & 3.63{\tiny$\pm0.53$} & 2.91{\tiny$\pm0.42$} & 3.53{\tiny$\pm0.55$} & 3.48 (87\%) & 3.52{\tiny$\pm0.61$} & 3.52{\tiny$\pm0.61$} & 3.52{\tiny$\pm0.61$} & 24.11{\tiny$\pm3.13$} & 4402.94 \\
        \midrule
        ChatLlama2-7B & 2.01{\tiny$\pm0.87$} & 1.31{\tiny$\pm0.65$} & 1.49{\tiny$\pm0.68$} & 0.72 (18\%) & 0.83{\tiny$\pm0.82$} & 0.85{\tiny$\pm0.79$} & 1.14{\tiny$\pm0.94$} & 8.34{\tiny$\pm4.50$} & 1763.87 \\
        CodeLlama2-7B & 2.04{\tiny$\pm0.91$} & 1.34{\tiny$\pm0.66$} & 1.64{\tiny$\pm0.71$} & 0.60 (15\%)& 0.99{\tiny$\pm0.90$} & 1.00{\tiny$\pm0.89$} & 1.22{\tiny$\pm1.00$} & 8.83{\tiny$\pm4.91$} & 1786.59\\
        ChatLlama2-13B & 1.97{\tiny$\pm0.88$} & 1.24{\tiny$\pm0.64$} & 1.53{\tiny$\pm0.69$} & 0.72 (18\%)& 1.02{\tiny$\pm0.84$} & 1.03{\tiny$\pm0.79$} & 1.24{\tiny$\pm0.89$} & 8.75{\tiny$\pm4.49$} & 2032.84\\
        ChatLlama2-70B & 2.32{\tiny$\pm0.90$} & 1.51{\tiny$\pm0.67$} & 1.78{\tiny$\pm0.73$} & 0.84 (21\%)& 1.25{\tiny$\pm0.92$} & 1.27{\tiny$\pm0.90$} & 1.60{\tiny$\pm0.96$} & 10.57{\tiny$\pm4.85$} & 2487.18\\
        \midrule
        Phi3Medium & 2.90{\tiny$\pm1.25$} & 1.74{\tiny$\pm0.81$} & 2.33{\tiny$\pm1.04$} & 1.64 (41\%) & 2.45{\tiny$\pm1.21$} & 2.33{\tiny$\pm1.14$} & 2.58{\tiny$\pm1.18$} & 15.97{\tiny$\pm6.50$} & 3314.35 \\
        Phi4 & 2.94{\tiny$\pm0.24$} & 2.55{\tiny$\pm0.58$} & 2.87{\tiny$\pm0.35$} & 2.16 (54\%) & 2.84{\tiny$\pm0.43$} & 2.82{\tiny$\pm0.45$} & 2.88{\tiny$\pm0.39$} & 19.06{\tiny$\pm2.26$} & 3815.88 \\

        CohereRPlus & 2.89{\tiny$\pm0.95$} & 1.70{\tiny$\pm0.64$} & 2.38{\tiny$\pm0.79$} & 1.68 (42\%) & 2.50{\tiny$\pm0.92$} & 2.42{\tiny$\pm0.86$} & 2.62{\tiny$\pm0.87$} & 16.20{\tiny$\pm4.86$} & 2851.20 \\
        \bottomrule
    \end{tabular}
    }
\end{table*}

GPT4O3Mini outperforms all others, achieving the highest total score of 24.11 out of 28. Within the GPT family, GPT4O1 incurs significantly higher token costs, up to twice those of GPT4O. GPT4O3Mini, a refined version, improves both efficiency and performance compared to GPT4O1. Models from the Llama family (from 7B to 70B), whether fine-tuned for code generation or not, exhibit relatively low scores. Generally, larger parameter scale enhances overall performance but also increases token usage.

\begin{tcolorbox}[colback=yy!30!white, colframe=rr]
\textbf{Finding 1}: LLMs often fail to produce complete and executable solutions, frequently overlooking steps like data pre-processing. While larger models handle import errors better, they still struggle with logic errors.
\end{tcolorbox}

LLMs often overlook essential procedures like data pre-processing (e.g., handling missing values) and filtering in the generated code, as revealed in \autoref{table:error-analysis}. For GPT4O3Mini, only 49\% of the generated code includes proper pre-processing and filtering, while the rest operate on unclean data. This issue is even more pronounced in smaller models. As shown in \autoref{table:error-analysis}, we observe the following:
(\romannumeral 1) Smaller models frequently face import errors, such as using non-imported functions, which decrease as model capacity grows.
(\romannumeral 2) Logic errors, including syntax issues and function call errors, persist across models, indicating the need for future improvements.
(\romannumeral 3) Fine-tuning on code helps CodeLlama2-7B reduce import errors but increases logic errors, resulting in a lower executable code rate than ChatLlama2-7B.

\begin{table}[htbp]
    \centering
    \caption{Analysis on the portion of the codes without pre-processing, and the frequencies of error types.}
    \label{table:error-analysis}
    \resizebox{.999\linewidth}{!}{
        \begin{tabular}{lrrrrr}
            \toprule
            LLM & No Pre-processing & Import Error & Logic Error\\
            \midrule
            GPT3.5Turbo&  71\% & 3.8\% & 43.2\%\\
            GPT4Turbo & 66\% & 1.3\% & 20.7\%\\
            GPT4O & 66\% & 0.4\% & 18.6\%\\
            GPT4O1 & 50\% & 0.3\% & 14.7\%\\
            GPT4O3Mini & 51\% & 0.3\% & 12.7\%\\
            ChatLlama2-7B & 92\% & 41.8\% & 40.1\%\\
            CodeLlama2-7B & 89\% & 38.2\% & 46.8\%\\
            ChatLlama2-13B & 87\% & 36.1\% & 45.9\%\\
            ChatLlama2-70B & 87\% & 15.0\% & 64.0\%\\
            Phi3Medium & 72\% & 5.9\% & 53.1\%\\
            Phi4 & 58\% & 3.8\% & 42.2\%\\
            CohereRPlus & 78\% & 4.6\%  & 53.4\%\\
            \bottomrule
        \end{tabular}
    }
\end{table}

\begin{tcolorbox}[colback=yy!30!white, colframe=rr]
\textbf{Finding 2}: Fine-tuning on code generation improves the overall performance of models, sometimes allowing them to perform better than their parameter size would suggest.
\end{tcolorbox}
With 46\% fewer parameters, CodeLlama2-7B outperforms ChatLlama2-13B in total scores, aligning with the conclusion of \citep{zhou2023solving}. Fine-tuning on code remarkably improves model performance in code relevance, code-text alignment, and the quality of textual description.

\begin{tcolorbox}[colback=yy!30!white, colframe=rr]
\textbf{Finding 3}: Fine-tuning for code generation increases the frequency and length of code produced by CodeLlama2-7B, but it \textbf{negatively impacts} the model's ability to generate executable data analysis code.
\end{tcolorbox}

To our surprise, CodeLlama2-7B achieves 15\% of the executable code rate, falling behind even ChatLlama2-7B. We find that, at this parameter level, fine-tuning on code generation helps to boost the performance of other domains, including textual analysis and text-code alignment. However, in terms of executable code rate, its performance is downgraded. This may be attributed to the overly specialized fine-tuning. Also, the reason may be that fine-tuning on code does not always focus on the data-analysis-related codes. For example, there may be some C++ kernel code that is not related to data analysis.

We systematically analyze the lengths and lines of the generated codes in \autoref{table:code-length}. Despite our prompt instructions to generate codes, we find that models from the Llama family often ignore code generation.
It turns out that fine-tuning on code only helps CodeLlama2-7B generate code more frequently, with a tendency to generate longer code. However, for a complex task like data-aware predictive analysis, it won't improve the quality of the generated code at this scale.

\begin{table}[htbp]
    \centering
    \caption{Analysis of the average length and number of lines of the generated code, as well as the ratio of cases with no code.}
    \label{table:code-length}
        \begin{tabular}{lrrrrr}
            \toprule
            LLM & Length & \#Lines & Absent\\
            \midrule
            GPT3.5Turbo&  967.45 & 27.22 & 4\%\\
            GPT4Turbo & 1276.49 & 32.46 & 0\\
            GPT4O & 2423.53 & 65.79 & 0\\
            GPT4O1 & 4375.41 & 127.37 & 0\\
            GPT4O3Mini & 3561.61 & 90.63 & 0\\
            ChatLlama2-7B & 660.72 & 17.41 & 64\%\\
            CodeLlama2-7B & 863.85 & 21.64 & 60\%\\
            ChatLlama2-13B & 627.51 & 16.68 & 63\%\\
            ChatLlama2-70B &  611.32 & 16.54 & 53\%\\
            Phi3Medium & 1336.94 & 30.79 & 0\\
            Phi4 & 1469.85 & 33.85 & 0\\
            CohereRPlus & 1749.12 & 52.90  & 0\\
            \bottomrule
        \end{tabular}
\end{table}

\begin{tcolorbox}[colback=yy!30!white, colframe=rr]
\textbf{Finding 4}: LLMs vary in score distribution across data source domains. GPT4O3Mini and GPT4O are well-balanced, while ChatLlama2-70B excels in the Education domain.
\end{tcolorbox}

\begin{figure}[h]
    \centering
    \begin{subfigure}{0.49\linewidth}
        \centering
        \includegraphics[width=\linewidth]{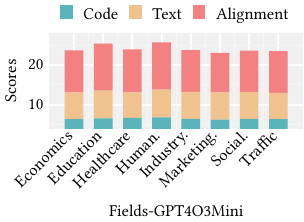}
    \end{subfigure}
    \begin{subfigure}{0.49\linewidth}
        \centering
        \includegraphics[width=\linewidth]{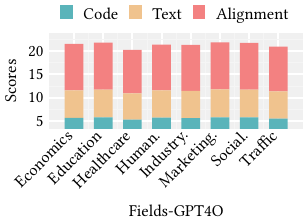}
    \end{subfigure}
    \begin{subfigure}{0.49\linewidth}
        \centering
        \includegraphics[width=\linewidth]{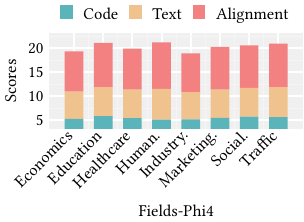}
    \end{subfigure}
    \begin{subfigure}{0.49\linewidth}
        \centering
        \includegraphics[width=\linewidth]{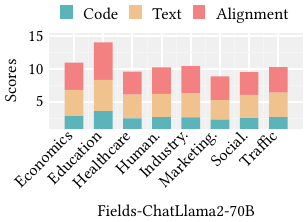}
    \end{subfigure}
    \caption{Score distributions of LLMs on eight fields. For clarity we present the total scores of text, code, and their alignment.}
    \label{fig:fields}
\end{figure}

We present the performance of leading models — GPT4O3Mini, GPT4O, Phi4, and ChatLlama2-70B — within their respective families across eight fields (see \autoref{fig:fields}). The data highlight the varying effectiveness of these LLMs across domains. GPT4O3Mini and GPT4O demonstrate balanced performance, with deviations of 1.67 and 1.82 points, respectively. In contrast, Phi4 and ChatLlama2-70B exhibit greater variability, with score differences of 4.38 and 5.16 points. The score distributions differ across fields. For instance, ChatLlama2-70B excels in \textit{Education}, exceeding its average score by 31.4\%, suggesting domain-specific strengths and weaknesses among models.

\subsection{Impact of Context Length}
\begin{figure}
    \centering
    \includegraphics[width=\linewidth]{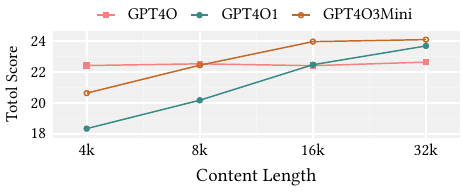}
    \caption{Analysis on impact of context length limit.}
    \label{fig:context-length}
\end{figure}

We evaluate the impact of context length on GPT4O, GPT4O1, and GPT4O3Mini in \autoref{fig:context-length}. At 4k tokens, GPT4O performs well, and increasing the token limit does not improve its performance. However, GPT4O1 and GPT4O3Mini, which need extra tokens for reasoning, perform much worse initially\footnote{As stated in \autoref{sec:settings}, we set the context length to 32,768 for GPT4O1 and GPT4O3Mini to unlock their full potential, resulting in the outcomes in \autoref{table:main-evaluation}.}. As context length increases, their performance improves rapidly, with GPT4O3Mini achieving similar results to GPT4O1 using fewer tokens.

\subsection{LLMs as Evaluators}

\begin{table}[htbp]
    \centering
    \small
    \caption{Results on PredictiQ with different evaluators.}
    \label{table:eval-by-other-llms-compact}
    \resizebox{.999\linewidth}{!}{
    \begin{tabular}{lcccc}
        \toprule
        Evaluator & GPT4Turbo & GPT4O & Phi3Medium  \\ 
        \midrule
        GPT3.5Turbo &  17.21{\tiny$\pm4.82$} & 21.68{\tiny$\pm3.69$} & 25.49{\tiny$\pm1.29$} \\
        GPT4Turbo &  20.34{\tiny$\pm3.92$} & 25.34{\tiny$\pm2.08$}  & 26.45{\tiny$\pm1.42$} \\
        GPT4O & 22.43{\tiny$\pm3.72$} & 26.13{\tiny$\pm1.97$} & 26.86{\tiny$\pm1.93$} \\
        GPT4O1 & 23.70{\tiny$\pm3.40$} & 26.45{\tiny$\pm2.33$} & 26.96{\tiny$\pm1.22$}\\
        GPT4O3Mini & 24.15{\tiny$\pm3.13$} & 26.64{\tiny$\pm2.05$} & 27.04{\tiny$\pm1.50$}\\
        ChatLlama2-7B & 8.34{\tiny$\pm4.50$} & 13.94{\tiny$\pm5.02$} & 22.93{\tiny$\pm4.56$} \\
        CodeLlama2-7B & 8.83{\tiny$\pm4.91$} & 14.59{\tiny$\pm4.89$} & 22.62{\tiny$\pm4.51$} \\
        ChatLlama2-13B & 8.75{\tiny$\pm4.49$} & 15.75{\tiny$\pm4.13$} & 23.67{\tiny$\pm3.34$}  \\
        ChatLlama2-70B & 10.57{\tiny$\pm4.85$} & 16.36{\tiny$\pm4.54$} & 23.54{\tiny$\pm4.13$}  \\
        Phi3Medium & 15.97{\tiny$\pm5.49$} & 21.19{\tiny$\pm4.67$} & 25.23{\tiny$\pm1.43$}  \\
        Phi4  & 19.06{\tiny$\pm2.26$} & 21.19{\tiny$\pm4.67$} & 25.23{\tiny$\pm1.43$}  \\
        CohereRPlus & 16.20{\tiny$\pm4.83$} & 20.87{\tiny$\pm4.54$}  & 25.07{\tiny$\pm1.28$} \\
        \bottomrule
    \end{tabular}
    }
\end{table}

\begin{figure}
    \centering
    \includegraphics[width=\linewidth]{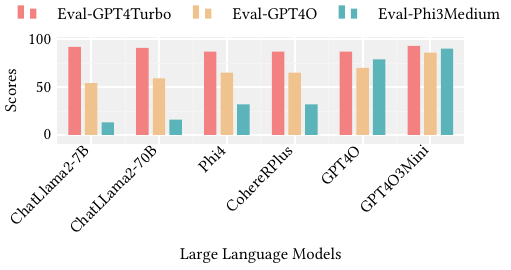}
    \caption{Alignment scores of different evaluators with human experts. See \autoref{fig:human-eval-full} for full results.}
    \label{fig:human-eval}
\end{figure}

To investigate how different LLM evaluators align with human preferences and select the proper evaluator, we engage (\romannumeral 1) five experts in data analysis and (\romannumeral 2) LLMs including GPT4Turbo, GPT4O, and Phi3Medium to grade the responses according to the evaluation protocol \footnote{We exclude GPT4O1 and GPT4O3Mini as evaluators due to their high computational cost and slow running speed.}. We then calculate the average scores from human experts and compare them with those from LLMs. 
Score pairs with an absolute difference smaller than 4 are labeled positive (\texttt{1}), while those with a larger difference are labeled negative (\texttt{0}). The alignment between LLM and expert scores is visualized in \autoref{fig:human-eval}, with higher scores indicating better alignment. Full results are presented in \autoref{fig:human-eval-full}.

\begin{tcolorbox}[colback=yy!30!white, colframe=rr]
\textbf{Finding 5}: GPT4Turbo's evaluations align most closely with those of the experts, while GPT4O tends to be more positively biased. Phi3Medium is not applicable because it lacks differentiation in its evaluations.
\end{tcolorbox}

As shown in \autoref{fig:human-eval}, GPT4Turbo aligns most closely with expert ratings, matching 90.5\% of instances and maintaining a score distribution consistent with human evaluators across all target models. In contrast, GPT4O shows weaker alignment, particularly with the Llama family, where it frequently assigns disproportionately high scores to poor responses. Phi3Medium demonstrates the poorest alignment, assigning nearly indistinguishable scores across all targets. This may be attributed to the \textit{round number bias} \cite{roundnumberbias,stureborg2024largelanguagemodelsinconsistent}, where certain scores are assigned more frequently, irrespective of the context, impeding precise quality assessments. This could result from biased training data, sentiment-supportive fine-tuning, or limitations in reasoning abilities \cite{murugadoss2024evaluatingevaluatormeasuringllms, wei2024systematicevaluationllmasajudgellm, wang-etal-2024-large-language-models-fair}.

\begin{tcolorbox}[colback=yy!30!white, colframe=rr]
\textbf{Finding 6}: The tendency of LLMs to assign themselves the highest scores when serving as evaluators is reduced in the context of predictive analysis.
\end{tcolorbox}
Studies \cite{ye2024justiceprejudicequantifyingbiases, panickssery2024llmevaluatorsrecognizefavor, koo-etal-2024-benchmarking} show that when LLMs are used as evaluators, they often exhibit a bias towards assigning higher scores to their own responses. However, in our experimental setup, this bias appears to be mitigated. 
As demonstrated in \autoref{table:eval-by-other-llms-compact}, the gap between total scores ($\text{Score}_\text{GPT4O}-\text{Score}_\text{GPT4Turbo}$), when evaluated by GPT4Turbo, is 2.09. In contrast, the gap narrows to 0.79 when evaluated by GPT4O. Interestingly, when the evaluator is switched from GPT4Turbo to GPT4O, the performance gap narrows rather than widens, indicating that the "egocentric" \cite{koo-etal-2024-benchmarking} or "self-preference" \cite{panickssery2024llmevaluatorsrecognizefavor} scoring tendency is reduced in this evaluation.

\subsection{Ablation Study}

\begin{table}[htbp]
    \centering
    \small
    \caption{Ablation study. For clarity, we present the total scores of three domains.}
    \label{table:ablation}
    \resizebox{.999\linewidth}{!}{
    \begin{tabular}{lcccc}
        \toprule
        & \textbf{Text} & \textbf{Code} & \textbf{Alignment} & \textbf{Total} \\ 
        \midrule
        GPT4O &  5.99{\tiny$\pm1.22$} & 6.36{\tiny$\pm0.67$} & 10.08{\tiny$\pm1.93$} & 22.43{\tiny$\pm3.82$} \\
        w/o Role Play & 5.78{\tiny$\pm1.14$} & 6.22{\tiny$\pm0.67$} & 9.94{\tiny$\pm2.07$} & 21.94{\tiny$\pm3.88$} \\ 
        w/o Data Summary & 5.81{\tiny$\pm1.12$} & 6.34{\tiny$\pm0.68$} & 9.76{\tiny$\pm2.17$} & 21.91{\tiny$\pm3.97$} \\
        \midrule
        GPT4O3Mini & 6.54{\tiny$\pm0.76$} & 7.01{\tiny$\pm0.55$} & 10.56{\tiny$\pm1.83$} & 24.11{\tiny$\pm3.13$} \\
        w/o Role Play & 6.56{\tiny$\pm0.86$} & 7.31{\tiny$\pm0.67$} & 10.35{\tiny$\pm1.90$} & 24.22{\tiny$\pm4.07$} \\
        w/o Data Summary & 6.25{\tiny$\pm0.56$} & 7.18{\tiny$\pm0.67$} & 10.22{\tiny$\pm1.76$} & 23.65{\tiny$\pm3.13$} \\
        \midrule
        ChatLlama2-70B & 3.98{\tiny$\pm1.46$} & 2.66{\tiny$\pm0.76$} & 4.39{\tiny$\pm2.64$} & 11.03{\tiny$\pm4.86$} \\
        w/o Role Play & 3.94{\tiny$\pm1.64$} & 2.65{\tiny$\pm0.82$} & 4.19{\tiny$\pm3.10$} & 10.78{\tiny$\pm5.56$} \\
        w/o Data Summary & 3.84{\tiny$\pm1.65$} & 2.64{\tiny$\pm0.77$} & 4.22{\tiny$\pm3.05$} & 10.70{\tiny$\pm5.46$} \\
        \midrule
        Phi4 & 5.49{\tiny$\pm0.70$} & 5.03{\tiny$\pm0.35$} & 8.54{\tiny$\pm1.21$} & 19.06{\tiny$\pm2.26$} \\
        w/o Role Play & 5.15{\tiny$\pm1.67$} & 4.45{\tiny$\pm1.54$} & 7.83{\tiny$\pm2.20$} & 17.61{\tiny$\pm2.46$} \\
        w/o Data Summary & 4.85{\tiny$\pm3.18$} & 4.68{\tiny$\pm2.47$} & 7.97{\tiny$\pm3.85$} & 17.50{\tiny$\pm3.15$} \\
        \midrule
        CohereRPlus & 4.59{\tiny$\pm1.50$} & 4.06{\tiny$\pm0.79$} & 7.54{\tiny$\pm2.57$} & 16.20{\tiny$\pm4.86$} \\
        w/o Role Play & 5.01{\tiny$\pm1.54$} & 4.21{\tiny$\pm0.80$} & 8.20{\tiny$\pm2.79$} & 17.42{\tiny$\pm5.13$} \\
        w/o Data Summary & 5.04{\tiny$\pm1.39$} & 4.21{\tiny$\pm0.72$} & 8.24{\tiny$\pm2.38$} & 17.49{\tiny$\pm4.50$} \\
        \bottomrule
    \end{tabular}
    }
\end{table}

\begin{tcolorbox}[colback=yy!30!white, colframe=rr]
\textbf{Finding 7}: Prompt engineering primarily influences textual descriptions and text-code alignment, with relatively little impact on code generation.
\end{tcolorbox}

We conduct ablation studies using a subset of 160 queries covering all fields from Table \ref{table:data-query-statistics}. We present the results of the top-performing models from each family in \autoref{table:ablation}, with the full results provided in \autoref{table:ablation-full}. The results show that prompt engineering primarily impacts textual descriptions and text-code alignment, with minimal effect on code generation. They enhance the performance of most models, except for CohereRPlus. We also find that prompt engineering has little impact on GPT4O3Mini, likely because the reasoning modules have similar functionalities, which is consistent with the findings in its report \footnote{https://platform.openai.com/docs/guides/reasoning}.

\subsection{LLMs for Predictive Analysis}

Based on the aforementioned evaluations, existing LLMs still have significant room for improvement in predictive analysis, particularly in terms of performance and efficiency. While GPT4O3Mini achieves relatively high scores, it lacks depth in providing justifications for algorithmic choices (Depth score: 2.91 out of 4) and struggles with data pre-processing (51\% of codes omit this step). Additionally, its high token consumption (with an average exceeding 4K tokens) and the requirement for a large context window (up to 32K tokens) present challenges for real-world applications, especially in privacy-sensitive scenarios where only local, smaller models are viable.
\section{Related Works}
\subsection{Large Language Models}
Recent advancements in natural language processing (NLP) are largely attributed to Large Language Models (LLMs) \citep{zhao2023survey,chang2023survey} like GPT series \citep{brown2020language,openai2023gpt4,openai2024gpt4ocard}, Gemini series \citep{geminiteam2024gemini15unlockingmultimodal}, Llama series \citep{touvron2023llama,CodeLlama,grattafiori2024llama3herdmodels}, and so on. These models undergo extensive pre-training on vast text datasets and are further refined through techniques like reinforcement learning from human feedback (RLHF) \citep{kirk2024understanding} and instruction fine-tuning (IFT) \citep{brown2020language}. 

In parallel, models specifically fine-tuned by code-formatted data emerged, considering the wide-ranging demand for code understanding and program generation from natural language prompts. For example, Codex \citep{chen2021evaluating}, CodeT \citep{Chen2022CodeTCG}, Code Llama \citep{CodeLlama} and HiRoPE \citep{zhang-etal-2024-hirope} have displayed excellent performance in code-related tasks, such as code completion and description-to-code generation. Meanwhile, these fine-tuned LLMs have shown extra progress in solving math problems \citep{zhou2023solving, xu-etal-2024-chatglm}.

\subsection{LLMs for data analysis}

LLMs have recently demonstrated significant potential in various data analysis fields. Some studies concentrate on rudimentary data tasks in preparation for further analysis like early-stage data preparation \citep{Zhang2023LargeLM} and table-to-text summarization \citep{zhao-etal-2023-investigating}. Some studies \citep{chen2022large,saeed2023querying,gao2023texttosql} focus on LLMs' performance in transforming texts into SQL queries with their execution on the corresponding database, mainly for elementary descriptive analysis. For more advanced data analysis, most existing research works \citep{yu2023harnessing,jin2023time,schoenegger2023large,pratt2024languagemodelsuseforecasting, hong2024datainterpreterllmagent} are limited to domain-specific time series forecasting tasks and lack generalizability. Text2Analysis \citep{he2023text2analysis} introduces queries for four types of advanced data analysis. However, their evaluation is exclusively centered on code generation, neglecting the essential elements of textual explanation and the alignment between text and code. Generally, there is a noticeable lack of comprehensive evaluations of LLMs in predictive analysis.
\section{Conclusion}
In this paper, we propose the PredictiQ benchmark, addressing the gap in evaluating LLMs' capabilities in predictive analysis. The benchmark provides domain-diversified datasets, data-specific queries, and in-depth evaluation protocols. We evaluate twelve popular LLMs on the benchmark, highlighting their capabilities and limitations in predictive analysis.
From the evaluation, we find the following key insights:
(i) Fine-tuning LLMs on code boosts their performance, sometimes exceeding the limits of model parameters.
(ii) Code generation and text analysis abilities are interconnected, jointly influencing LLMs' predictive analysis skills.
(iii) LLMs exhibit varying levels of predictive ability across domains.
Overall, existing LLMs still have significant room for improvement in achieving both high performance and efficiency in predictive analysis.
\section{Limitations}
This work exhibits the following limitations. 
(\romannumeral 1) As discussed in \autoref{sec:Problem Definition}, this work centers on evaluating the performance of LLMs in predictive analysis, and does not extend to other advanced data analysis fields such as prescriptive analysis or diagnostic analysis. Future research is expected to investigate broader analysis dimensions and assess effectiveness of LLMs across diverse analytical domains.
(\romannumeral 2) The dataset employed in this study is confined to a limited number of commonly encountered fields, potentially overlooking requirements and challenges present in less-represented or novel fields. This limitation may impact the generalizability of the findings and suggests that future work should incorporate more diverse datasets to explore how LLMs perform across a broader range of fields.

\section{Ethics Statement}
This work benchmarks the efficacy of predictive analysis in LLMs. The evaluation involves datasets collected from publicly available platforms, detailed in \autoref{sec:appendix_datasets}. 
We ensure that the collected datasets do not contain any personally identifiable information. Additionally, we rigorously verify the data licenses and copyright permissions to confirm that they authorize public use for research purposes. 
We need to further clarify that this study focuses on the evaluation of language models and does not introduce new model architectures or deployment strategies. While we acknowledge that flawed model predictions, particularly in sensitive or high-stakes domains, can contribute to biased or harmful outcomes, addressing such risks requires domain-specific considerations and interventions that fall beyond the scope of this work. The evaluation intends to inform and support future research aimed at improving model reliability and mitigating downstream risks in applied settings.

In the study, volunteers consisting of master's students in data science with an Asian background conducted human annotation to generate and evaluate the queries. They also participate in manual code execution and LLM response evaluation.
While these annotators possess a solid foundation of predictive analysis, there is a potential risk that individuals from a specific cultural background may exhibit biases in their comprehension of query content and LLM responses.

We have used ChatGPT to assist us in refining the expression of our paper.

\bibliography{coling_latex}

\appendix

\section{Datasets}
\label{sec:appendix_datasets}

\begin{table*}[htbp]
\centering
\caption{Detailed statistics of collected datasets and the corresponding queries. }
\label{table:table-statistics}
\begin{tabular}{llccc}
\toprule
\textbf{Table Name}  & \textbf{Domain} & \textbf{\#Columns} & \textbf{\#Queries} \\
\midrule

CarSales  & Marketing and Sales & 7 & 20 \\ 

MonetPaintingsSales  & Marketing and Sales & 6 & 20 \\ 

MovieBuzz  & Marketing and Sales & 13 & 20 \\ 

StoreSales  & Marketing and Sales & 6 & 20 \\ 

HousePrices  & Marketing and Sales & 81 & 100 \\ 

GasolineConsumption  & Marketing and Sales & 6 &  20 \\ 

GPA  & Education & 6 &  20 \\ 

ProgramEffectiveness  & Education & 5 & 20 \\ 

GenderEconomicsCourses  & Education & 11 & 30 \\ 

RunLog  & Healthcare & 5 &  20 \\ 

Smoker  & Healthcare & 24 & 40 \\ 

HealthCareOutcomes  & Healthcare & 16 &  30 \\ 

GermanHealthCare  & Healthcare & 25 &  40 \\ 

Employee  & Human Resource & 14 &  20 \\ 

LaborSupply  & Human Resource & 19 & 40 \\ 

LaborMarket  & Human Resource & 12 & 20 \\ 

StockPrice  & Economics & 9 &  20 \\ 

USD\_ISK\_Exchange\_Rate  & Economics & 6 & 20 \\ 

CostFunction  & Economics & 10 & 20 \\ 

InvestmentData  & Economics & 8 & 20 \\ 

Macroeconomics  & Economics & 14 & 20 \\ 

LongleyData  & Economics & 5 & 20 \\ 

ExpenditureAndDefault  & Economics & 14 &  30 \\ 

IncomeAndExpenditure  & Economics & 7 & 20 \\ 

MunnellProductivity  & Economics & 11 & 20 \\ 

KleinModel  & Economics & 10 & 20 \\ 

MunicipalExpenditure  & Economics & 5 & 20 \\ 

LaLondeEarnings  & Economics & 15 & 40 \\ 

Titanic  & Social Study & 12 &  20 \\ 

ExtramaritalAffairs  & Social Study & 11 &  30 \\ 

CrossExtramaritalAffairs  & Social Study & 10 & 30 \\ 

ShipAccidents  & Social Study & 14 & 30 \\ 

HourlyTrafficVolume  & Traffic & 9 &  20 \\ 

AirPassengerTraffic  & Traffic & 6 &  20 \\ 

CostData\_For\_US\_Airlines  & Traffic & 6 & 20 \\ 

TransportationEquipmentManufacturing  & Traffic & 5 & 20 \\ 

SwissRailroads  & Traffic & 21 &  50 \\ 

MoleFractionOfCarbonDioxide  & Industry Analysis & 8 & 20 \\ 

ShanghaiLicensePlatePrice  & Industry Analysis & 5 & 20 \\ 

GasolineMarket  & Industry Analysis & 11 & 20 \\ 

BaseballAttendance  & Industry Analysis & 7 & 20 \\ 

SpanishDairyFarmProduction  & Industry Analysis & 29 & 50 \\ 

TravelModeChoice  & Industry Analysis & 7 &  20 \\ 

CaliforniaUtilities  & Industry Analysis & 10 &  30 \\ 

\bottomrule
\end{tabular}
\end{table*}

\begin{table}[h!]
\centering
\begin{tabular}{lc}
\toprule
\textbf{Type} & \textbf{\#Queries} \\
\midrule
Classification & 285 \\
Regression & 324 \\
Forecasting & 220 \\
Clustering & 137 \\
Anomaly Detection & 126 \\
Other & 38 \\
\bottomrule
\end{tabular}
\caption{Detailed statistics of query task categories.}
\label{table:query_type}
\end{table}

We have collected tabular datasets from different application scenarios to ensure the benchmark's diversity and validity. The data sources can be summarized as follows.
\begin{itemize}
    \item Datasets of the \textbf{Economics} domain cover financial systems at a state or national level. For example, the yearly records of Gross Domestic Product (GDP), Gross National Product (GNP), and other economic metrics.
    \item Datasets of the \textbf{Marketing and Sales} domain cover sales records and marketing schemes for specific brands or products. This includes, for instance, annual data on costs, sales, and profits for various car models across multiple brands.
    \item Datasets of the \textbf{Industry Analysis} domain cover the development of a certain industry. For example, they might include data on the distribution of residential electricity demands within the utility sector.
    \item Datasets of the \textbf{Traffic} domain cover different types of traffic records. For example, the monthly records of passenger volumes in several airports. 
    \item Datasets of the \textbf{Healthcare} domain cover healthcare records. For example, the records of physical examination indexes for both smokers and non-smokers. 
    \item Datasets of the \textbf{Social Study} domain cover specific sociological issues. For example, the records of demographic and personal variables as well as the occurrence of extramarital affairs. 
    \item Datasets in the \textbf{Human Resource} domain encompass aspects of human resource management, such as detailed records of employee information, including salary, department, and other related data.
    \item Datasets in the \textbf{Education} domain include records related to educational activities, such as college students' grades and attendance data.
\end{itemize}

Datasets are acquired from various public platforms, including the data science platform Kaggle \footnote{https://www.kaggle.com/}, the TCPD benchmark for time-series analysis \citep{TCPD} and the eighth edition of Econometric Analysis \citep{greene2019econometric}. 
We ensure that the collected datasets do not contain personally identifiable information. Additionally, we verify the data licenses to ensure they are permitted for public use in research activities.
This guarantees that each dataset has undergone a thorough privacy and security review process. 
Details regarding the collected tabular datasets are available in \autoref{table:table-statistics}. 
Meanwhile, PredictiQ encompasses a wide range of tasks, including classification, regression, forecasting, clustering, and anomaly detection. We also provide statistics that underscore the diversity of our queries in \autoref{table:query_type}.

\section{Evaluation Protocols}

We provide a systematical description of our evaluation protocol in \autoref{table:evaluation} regarding text analysis, code generation, and text-code alignment. \textbf{Functional Correctness} is evaluated through the ratio of code execution without errors.

\begin{table*}[h]
    \centering
    \small
    \caption{The evaluation protocol. Relevance and Depth for text analysis. Usefulness and Functional Correctness correspond to code generation. Descriptive Accuracy, Coverage, and Clarity pertain to text-code alignment.}
    \label{table:evaluation}
    \resizebox{.999\linewidth}{!}{
    \begin{tabular}{p{1.5cm}p{1cm}p{14.5cm}}
        \toprule
        & \textbf{Score} & \textbf{Description} \\
        \midrule
        \multirow{5}{*}{\textbf{Relevance}} 
        & 0 & The analysis is not at all relevant to the query. It does not address the core issue or utilize the data appropriately.\\
        & 1 & The analysis is slightly relevant to the query, with superfluous information or focusing on secondary issues. \\
        & 2 & The analysis is relevant, touching upon the central query and making use of the data but not thoroughly enough. \\
        & 3 & The analysis is generally relevant, closely aligning with the given data and query, albeit with minor deviations.\\
        & 4 & The analysis is highly relevant, directly addressing the question with appropriate and substantial use of the data. \\
        \midrule
        \multirow{5}{*}{\textbf{Depth}}  
        & 0 & The analysis does not provide any justification for the choice of algorithm used. \\
        & 1 & The analysis provides minimal justification for the choice, with surface-level reasoning without delving into specifics. \\
        & 2 & The analysis moderately justifies the choice, with reasoned arguments but not fully covering all relevant factors. \\
        & 3 & The analysis gives a detailed justification. It is well-reasoned with minor gaps in the argumentation. \\
        & 4 & The analysis thoroughly justifies the choice, showing deep understanding and covering all aspects in detail. \\
        \midrule
        \multirow{5}{*}{\textbf{Usefulness}} 
        & 0 & The code is not at all helpful, it is irrelevant to the problem.\\
        & 1 & The code is slightly helpful, it contains information relevant to the problem, but it is easier to rewrite the solution. \\
        & 2 & The code is somewhat helpful, it requires significant changes (compared to the size of the snippet), but is still useful. \\
        & 3 & The code is helpful but needs to be slightly changed to solve the problem.\\
        & 4 & The code is very helpful, it solves the problem. \\
        \midrule
        \multirow{2}{*}{\makecell[l]{\textbf{Functional} \\ \textbf{Correctness}}} & \multirow{2}{*}{0-4}  & \multirow{2}{*}{Functional Correctness is evaluated through the ratio of code execution without errors, scaled up to 0-4.}   \\  \\
        \midrule
        \multirow{5}{*}{\makecell[l]{\textbf{Descriptive} \\ \textbf{Accuracy}}} 
        & 0 & The text does not describe the code at all; there is a complete mismatch. \\ 
        & 1 & The text describes less than half of what is represented in the code, missing significant functionalities. \\
        & 2 & The text describes about half of the code, yet there are still substantial gaps in the description. \\
        & 3 & The text describes most of the code adequately, with only minor gaps or discrepancies. \\
        & 4 & The text provides an accurate description of the code, perfectly aligning with all functionalities in the code snippet. \\
        \midrule
        \multirow{5}{*}{\textbf{Coverage}}  
        & 0 & The text does not cover any aspects of the code; the explanation is absent or irrelevant. \\
        & 1 & The text covers a few aspects of the code but misses out on explaining several crucial functionalities. \\
        & 2 & The text covers around half of the code's aspects, providing a partial understanding of the code's functionalities. \\
        & 3 & The text covers most of the aspects, offering a substantial understanding, with minor aspects left uncovered. \\
        & 4 & The text comprehensively covers all aspects of the code, leaving no functionalities unexplained. \\
        \midrule
        \multirow{5}{*}{\textbf{Clarity}} 
        & 0 & The alignment between the text and the code is unclear, making it difficult for users to follow.\\
        & 1 & The alignment is slightly clear; however, users may struggle to correlate the text and code without efforts.\\
        & 2 & The alignment is moderately clear, allowing users to somewhat follow the explanation with a reasonable effort. \\
        & 3 & The alignment is largely clear, aiding users in following the explanation with minimal difficulties. \\
        & 4 & The alignment is crystal clear, offering users an effortless pathway to understand the code through the text. \\
        \bottomrule
    \end{tabular}
    }
\end{table*}

\section{Settings}

\label{sec:settings}

For all LLMs, we set the temperature to 0.7 (where applicable), top\_p to 0.95 (where applicable), and a maximum token limit of 4096. For GPT4O1 and GPT4O3Mini, which require extra tokens for reasoning, we extend the token limit to 32,768.
\section{Additional Experiments}

\subsection{Quality Control During Annotation}

To further evaluate the reliability of expert annotations, we report Krippendorff’s Alpha coefficients in \autoref{table:dimension-scores}, which measure inter-annotator agreement across evaluation dimensions. The results indicate a generally substantial level of consistency among the annotators, supporting the overall quality and reliability of the annotation process.

\begin{table}[htbp]
\centering
\caption{Krippendorff’s Alpha coefficients showing inter-annotator agreement across evaluation dimensions.}
\label{table:dimension-scores}
\resizebox{.999\linewidth}{!}{
\begin{tabular}{lc}
\toprule
 & Krippendorff’s Alpha coefficient \\
\midrule
Relevance            & 0.89 \\
Depth                & 0.76 \\
Usefulness           & 0.84 \\
Descriptive Accuracy & 0.93 \\
Coverage             & 0.91 \\
Clarity              & 0.83 \\
\bottomrule
\end{tabular}}
\end{table}

\subsection{More Detailed Error Analysis}

Here We introduce more detailed error categories, including syntax, runtime, library, and data flow errors, to provide clearer insights into the model's performance and potential failure modes, detailed in \autoref{table:error-analysis-detailed}. Our analysis shows that increasing model capacity significantly reduces syntax errors. Moreover, fine-tuning on code further reduces syntax errors, although it tends to increase runtime and data flow errors.

\begin{table*}[htbp]
    \centering
    \caption{Error analysis across models. The top-level columns include the percentage of code without pre-processing and import errors. The logic error category is further broken down into subtypes of syntax, runtime, library, and data flow errors.}
    \label{table:error-analysis-detailed}
    \resizebox{.999\linewidth}{!}{
        \begin{tabular}{lcccccc}
            \toprule
            \multirow{2}{*}{LLM} & \multirow{2}{*}{No Pre-processing} & \multirow{2}{*}{Import Error} & \multicolumn{4}{c}{Logic Error} \\
            \cmidrule(lr){4-7}
            & & & Syntax Error & Runtime Error & Library Error & Data Flow Error \\
            \midrule
            GPT3.5Turbo      & 71\% & 3.8\%  & 15.3\% & 7.6\%  & 7.8\%  & 12.5\% \\
            GPT4Turbo        & 66\% & 1.3\%  & 7.8\%  & 3.8\%  & 3.4\%  & 5.7\% \\
            GPT4O            & 66\% & 0.4\%  & 5.1\%  & 3.9\%  & 3.8\%  & 5.8\% \\
            GPT4O1           & 50\% & 0.3\%  & 3.1\%  & 3.2\%  & 3.5\%  & 4.9\% \\
            GPT4O3Mini       & 51\% & 0.3\%  & 3.1\%  & 2.3\%  & 2.9\%  & 4.4\% \\
            ChatLlama2-7B    & 92\% & 41.8\% & 19.3\% & 8.3\%  & 6.2\%  & 6.3\% \\
            CodeLlama2-7B    & 89\% & 38.2\% & 17.4\% & 11.3\% & 8.8\%  & 9.3\% \\
            ChatLlama2-13B   & 87\% & 36.1\% & 18.3\% & 13.7\% & 8.6\%  & 5.3\% \\
            ChatLlama2-70B   & 87\% & 15.0\% & 12.5\% & 22.7\% & 16.3\% & 12.5\% \\
            Phi3Medium       & 72\% & 5.9\%  & 15.7\% & 12.4\% & 13.7\% & 11.3\% \\
            Phi4             & 58\% & 3.8\%  & 12.3\% & 10.3\% & 9.5\%  & 10.2\% \\
            CohereRPlus      & 78\% & 4.6\%  & 18.0\% & 10.4\% & 12.3\% & 12.7\% \\
            \bottomrule
        \end{tabular}
    }
\end{table*}

\subsection{Evaluation with Other LLMs}

We provide the detailed results adopting GPT4O and Phi3Medium as evaluators in \autoref{table:main-evaluation-gpt4o} and \autoref{table:main-evaluation-phi3medium}. We find that GPT4O is far more positive when conducting evaluation, compared to GPT4Turbo. Meanwhile, Phi3Medium is non-applicable as a rationale evaluator, as it tends to award high scores to almost all submissions, regardless of evident shortcomings. For example, as we have evidenced in \autoref{table:code-length}, many instances from Llama family models have no code at all, while Phi3Medium still scores them with high scores in code and text-code alignment domains. This makes the evaluation by Phi3Medium less meaningful. 

We have additionally incorporated Gemini-2.0-Flash \footnote{\url{https://deepmind.google/models/gemini/flash/}} as an alternative evaluator in our study. The corresponding results, presented in \autoref{table:main-evaluation-gemini}, are generally slightly more positive than those from GPT-4 Turbo. We further provide the weighted Cohen’s Kappa scores in \autoref{table:evaluation-kappa} to assess inter-rater agreement among GPT-4 Turbo, Gemini-2-Flash, and the human experts, with multiple expert evaluations consolidated into unified scores through majority voting. The scores indicate that Gemini-2-Flash generally also exhibits strong agreement with human experts, particularly regarding the Usefulness metric. Based on these findings, it is reasonable to adopt Gemini-2-Flash as the alternative evaluator.

\begin{table*}[htbp]
    \centering
    \caption{Evaluation results on PredictiQ on seven perspectives, each ranked from $0$ to $4$ following protocols in \autoref{table:evaluation}. For \textit{Functional Correctness} we measure the ratio of executable codes, and map it into scores from $0$ to $4$ evenly. We adopt the GPT4O as the evaluator.}
    \label{table:main-evaluation-gpt4o}
    \resizebox{.999\linewidth}{!}{
    \begin{tabular}{lccccccccc}
        \toprule
        & \multicolumn{2}{c}{\textbf{Text}} & \multicolumn{2}{c}{\textbf{Code}} & \multicolumn{3}{c}{\textbf{Text-Code Alignment}} & \multirow{3}{*}{\textbf{Total}} \\
        \cmidrule(lr){2-3}\cmidrule(lr){4-5}\cmidrule(lr){6-8}
         \multirow{2}{*}{LLM} & \multirow{2}{*}{Relevance} &  \multirow{2}{*}{Depth} &  \multirow{2}{*}{Usefulness} & Functional & Descriptive &  \multirow{2}{*}{Coverage} & \multirow{2}{*}{Clarity} &  \\ 
         & &  &  & Correctness & Accuracy &  & \\
         \midrule
            GPT3.5Turbo & 3.60{\tiny$\pm0.57$} & 2.62{\tiny$\pm0.72$} & 3.36{\tiny$\pm0.77$} & 2.12 (53\%) & 3.36{\tiny$\pm0.66$} & 3.13{\tiny$\pm0.66$} & 3.49{\tiny$\pm0.56$} & 21.68{\tiny$\pm3.69$} \\
            GPT4Turbo & 3.91{\tiny$\pm0.29$} & 3.24{\tiny$\pm0.50$} & 3.66{\tiny$\pm0.52$} & 3.12  (78\%)& 3.88{\tiny$\pm0.33$} & 3.63{\tiny$\pm0.49$} & 3.91{\tiny$\pm0.29$} & 25.34{\tiny$\pm2.08$} \\
            GPT4O & 3.95{\tiny$\pm0.26$} & 3.48{\tiny$\pm0.56$} & 3.83{\tiny$\pm0.40$} & 3.24 (81\%) & 3.90{\tiny$\pm0.30$} & 3.82{\tiny$\pm0.39$} & 3.91{\tiny$\pm0.29$} & 26.13{\tiny$\pm1.97$} \\
            GPT4O1 & 3.98{\tiny$\pm0.09$} & 3.66{\tiny$\pm0.27$} & 3.87{\tiny$\pm0.69$} & 3.24 (81\%) & 3.94{\tiny$\pm0.34$} & 3.85{\tiny$\pm0.45$} & 3.91{\tiny$\pm0.63$} & 26.45{\tiny$\pm2.33$} \\
            GPT4O3Mini & 3.99{\tiny$\pm0.06$} & 3.64{\tiny$\pm0.34$} & 3.89{\tiny$\pm0.18$} & 3.24 (81\%) & 3.94{\tiny$\pm0.07$} & 3.89{\tiny$\pm0.14$} & 3.96{\tiny$\pm0.09$} & 26.64{\tiny$\pm2.05$} \\
            ChatLlama2-7B & 2.74{\tiny$\pm0.81$} & 1.93{\tiny$\pm0.84$} & 2.27{\tiny$\pm0.90$} & 0.72 (18\%)& 1.96{\tiny$\pm0.96$} & 1.99{\tiny$\pm0.90$} & 2.33{\tiny$\pm0.89$} & 13.94{\tiny$\pm5.02$} \\
            CodeLlama2-7B & 2.79{\tiny$\pm0.81$} & 1.99{\tiny$\pm0.76$} & 2.31{\tiny$\pm0.92$} & 0.60 (15\%)& 2.19{\tiny$\pm0.95$} & 2.18{\tiny$\pm0.85$} & 2.53{\tiny$\pm0.83$} & 14.59{\tiny$\pm4.89$} \\
            ChatLlama2-13B & 3.03{\tiny$\pm0.63$} & 2.18{\tiny$\pm0.64$} & 2.63{\tiny$\pm0.79$} & 0.72 (18\%)& 2.31{\tiny$\pm0.84$} & 2.32{\tiny$\pm0.71$} & 2.56{\tiny$\pm0.76$} & 15.75{\tiny$\pm4.13$} \\
            ChatLlama2-70B & 3.09{\tiny$\pm0.79$} & 2.24{\tiny$\pm0.73$} & 2.66{\tiny$\pm0.89$} & 0.84 (21\%)& 2.35{\tiny$\pm0.86$} & 2.45{\tiny$\pm0.81$} & 2.73{\tiny$\pm0.75$} & 16.36{\tiny$\pm4.54$} \\
            Phi3Medium & 3.55{\tiny$\pm0.63$} & 2.74{\tiny$\pm0.71$} & 3.44{\tiny$\pm0.73$} & 1.64 (41\%)& 3.32{\tiny$\pm0.94$} & 3.14{\tiny$\pm0.91$} & 3.36{\tiny$\pm0.92$} & 21.19{\tiny$\pm4.67$} \\
            CohereRPlus & 3.53{\tiny$\pm0.70$} & 2.77{\tiny$\pm0.72$} & 3.20{\tiny$\pm0.86$} & 1.68 (42\%)& 3.22{\tiny$\pm0.87$} & 3.11{\tiny$\pm0.83$} & 3.36{\tiny$\pm0.73$} & 20.87{\tiny$\pm4.54$} \\
        \bottomrule
    \end{tabular}
    }
\end{table*}

\begin{table*}[htbp]
    \centering
    \caption{Evaluation results on PredictiQ on seven perspectives, each ranked from $0$ to $4$ following protocols in \autoref{table:evaluation}. For \textit{Functional Correctness} we measure the ratio of executable codes, and map it into scores from $0$ to $4$ evenly. We adopt the Phi3Medium as the evaluator.}
    \label{table:main-evaluation-phi3medium}
    \resizebox{.999\linewidth}{!}{
    \begin{tabular}{lccccccccc}
        \toprule
        & \multicolumn{2}{c}{\textbf{Text}} & \multicolumn{2}{c}{\textbf{Code}} & \multicolumn{3}{c}{\textbf{Text-Code Alignment}} & \multirow{3}{*}{\textbf{Total}} \\
        \cmidrule(lr){2-3}\cmidrule(lr){4-5}\cmidrule(lr){6-8}
         \multirow{2}{*}{LLM} & \multirow{2}{*}{Relevance} &  \multirow{2}{*}{Depth} &  \multirow{2}{*}{Usefulness} & Functional & Descriptive &  \multirow{2}{*}{Coverage} & \multirow{2}{*}{Clarity} &  \\ 
         & &  &  & Correctness & Accuracy &  & \\
         \midrule
            GPT3.5Turbo & 3.97{\tiny$\pm0.18$} & 3.60{\tiny$\pm0.54$} & 3.87{\tiny$\pm0.33$} & 2.12 (53\%)& 3.98{\tiny$\pm0.15$} & 3.95{\tiny$\pm0.21$} & 4.00 & 25.49{\tiny$\pm1.29$} \\
            GPT4Turbo & 3.96{\tiny$\pm0.20$} & 3.62{\tiny$\pm0.53$} & 3.84{\tiny$\pm0.42$} & 3.12 (78\%)& 3.97{\tiny$\pm0.18$} & 3.95{\tiny$\pm0.23$} & 4.00 & 26.45{\tiny$\pm1.42$} \\
            GPT4O & 3.99{\tiny$\pm0.11$} & 3.85{\tiny$\pm0.36$} & 3.98{\tiny$\pm0.15$} & 3.24 (81\%)& 3.90{\tiny$\pm0.61$} & 3.95{\tiny$\pm0.44$} & 3.95{\tiny$\pm0.44$} & 26.86{\tiny$\pm1.93$} \\
            GPT4O1 & 3.99{\tiny$\pm0.11$} & 3.95{\tiny$\pm0.36$} & 3.98{\tiny$\pm0.15$} & 3.24 (81\%)& 3.88{\tiny$\pm0.61$} & 3.97{\tiny$\pm0.44$} & 3.95{\tiny$\pm0.44$} & 26.96{\tiny$\pm1.22$} \\
            GPT4O3Mini & 3.99{\tiny$\pm0.08$} & 3.98{\tiny$\pm0.19$} & 3.98{\tiny$\pm0.10$} & 3.24 (81\%)& 3.95{\tiny$\pm0.61$} & 3.95{\tiny$\pm0.44$} & 3.95{\tiny$\pm0.44$} & 27.04{\tiny$\pm1.50$} \\
            ChatLlama2-7B & 3.83{\tiny$\pm0.69$} & 3.43{\tiny$\pm0.95$} & 3.68{\tiny$\pm0.76$} & 0.72 (18\%)& 3.69{\tiny$\pm0.96$} & 3.67{\tiny$\pm0.97$} & 3.91{\tiny$\pm0.48$} & 22.93{\tiny$\pm4.56$} \\
            CodeLlama2-7B & 3.84{\tiny$\pm0.58$} & 3.41{\tiny$\pm1.01$} & 3.54{\tiny$\pm0.81$} & 0.60 (15\%)& 3.67{\tiny$\pm0.89$} & 3.69{\tiny$\pm0.89$} & 3.88{\tiny$\pm0.60$} & 22.62{\tiny$\pm4.51$} \\
            ChatLlama2-13B & 3.92{\tiny$\pm0.47$} & 3.71{\tiny$\pm0.64$} & 3.82{\tiny$\pm0.55$} & 0.72 (18\%)& 3.80{\tiny$\pm0.79$} & 3.77{\tiny$\pm0.72$} & 3.92{\tiny$\pm0.47$} & 23.67{\tiny$\pm3.34$} \\
            ChatLlama2-70B & 3.85{\tiny$\pm0.60$} & 3.67{\tiny$\pm0.79$} & 3.71{\tiny$\pm0.73$} & 0.84 (21\%)& 3.80{\tiny$\pm0.85$} & 3.77{\tiny$\pm0.79$} & 3.90{\tiny$\pm0.61$} & 23.54{\tiny$\pm4.13$} \\
            Phi3Medium & 3.99{\tiny$\pm0.05$} & 3.81{\tiny$\pm0.39$} & 3.94{\tiny$\pm0.24$} & 1.64 (41\%)& 3.95{\tiny$\pm0.26$} & 3.94{\tiny$\pm0.28$} & 3.95{\tiny$\pm0.26$} & 25.23{\tiny$\pm1.43$} \\
            CohereRPlus & 3.98{\tiny$\pm0.15$} & 3.59{\tiny$\pm0.59$} & 3.88{\tiny$\pm0.33$} & 1.68 (42\%)& 3.98{\tiny$\pm0.15$} & 3.96{\tiny$\pm0.19$} & 4.00 & 25.07{\tiny$\pm1.28$} \\
            
        \bottomrule
    \end{tabular}
    }
\end{table*}

\begin{table*}[htbp]
    \centering
    \caption{Evaluation results on PredictiQ on seven perspectives, each ranked from $0$ to $4$ following protocols in \autoref{table:evaluation}. For \textit{Functional Correctness} we measure the ratio of executable codes, and map it into scores from $0$ to $4$ evenly. We adopt the Gemini-2.0-Flash as the evaluator.}
    \label{table:main-evaluation-gemini}
    \resizebox{.999\linewidth}{!}{
    \begin{tabular}{lccccccccc}
        \toprule
        & \multicolumn{2}{c}{\textbf{Text}} & \multicolumn{2}{c}{\textbf{Code}} & \multicolumn{3}{c}{\textbf{Text-Code Alignment}} & \multirow{3}{*}{\textbf{Total}} \\
        \cmidrule(lr){2-3}\cmidrule(lr){4-5}\cmidrule(lr){6-8}
         \multirow{2}{*}{LLM} & \multirow{2}{*}{Relevance} &  \multirow{2}{*}{Depth} &  \multirow{2}{*}{Usefulness} & Functional & Descriptive &  \multirow{2}{*}{Coverage} & \multirow{2}{*}{Clarity} &  \\ 
         & &  &  & Correctness & Accuracy &  & \\
         \midrule
            GPT3.5Turbo & 3.13 & 1.83 & 2.63 & 2.12 (53\%) & 2.83 & 2.65 & 2.9 & 18.09\\
            GPT4Turbo & 3.55 & 2.34 & 2.94 & 3.12  (78\%) & 3.14 & 3.12 & 3.3 & 21.51\\
            GPT4O & 3.75 & 2.65 & 3.45 & 3.24 (81\%) & 3.56 & 3.48 & 3.49 & 23.62\\
            GPT4O1 & 3.88 & 2.9 & 3.62 & 3.24 (81\%) & 3.66 & 3.65 & 3.55 & 24.66\\
            GPT4O3Mini & 3.93 & 3.01 & 3.61 & 3.24 (81\%) & 3.67 & 3.67 & 3.59 & 24.96\\
            ChatLlama2-7B & 2.31 & 1.65 & 1.58 & 0.72 (18\%) & 1.32 & 1.25 & 1.43 & 10.26\\
            CodeLlama2-7B & 2.33 & 1.77 & 1.76 & 0.60 (15\%) & 1.42 & 1.42 & 1.61 & 10.91\\
            ChatLlama2-13B & 2.18 & 1.75 & 1.66 & 0.72 (18\%) & 1.53 & 1.42 & 1.69 & 10.95\\
            ChatLlama2-70B & 2.76 & 1.83 & 1.75 & 0.84 (21\%) & 1.64 & 1.74 & 1.87 & 12.43\\
            Phi3Medium & 3.05 & 1.98 & 2.45 & 1.64 (41\%) & 2.67 & 2.7 & 2.76 & 17.25\\
            CohereRPlus & 3.06 & 1.87 & 2.56 & 1.68 (42\%) & 2.76 & 2.84 & 2.88 & 17.65\\

        \bottomrule
    \end{tabular}
    }
\end{table*}

\begin{table}[htbp]
\centering
\resizebox{.999\linewidth}{!}{
\begin{tabular}{lcc}
\toprule
 & GPT-4 Turbo vs Human & Gemini vs Human \\
\midrule
Relevance            & 0.85 & 0.74 \\
Depth                & 0.74 & 0.72 \\
Usefulness           & 0.78 & 0.86 \\
Descriptive Accuracy & 0.72 & 0.69 \\
Coverage             & 0.91 & 0.88 \\
Clarity              & 0.88 & 0.75 \\
\bottomrule
\end{tabular}}
\caption{Cohen’s Kappa scores assessing inter-rater agreement across GPT-4 Turbo, Gemini-2 Flash, and human experts on six evaluation dimensions.}
\label{table:evaluation-kappa}
\end{table}

\begin{figure*}
    \centering
    \includegraphics[width=\linewidth]{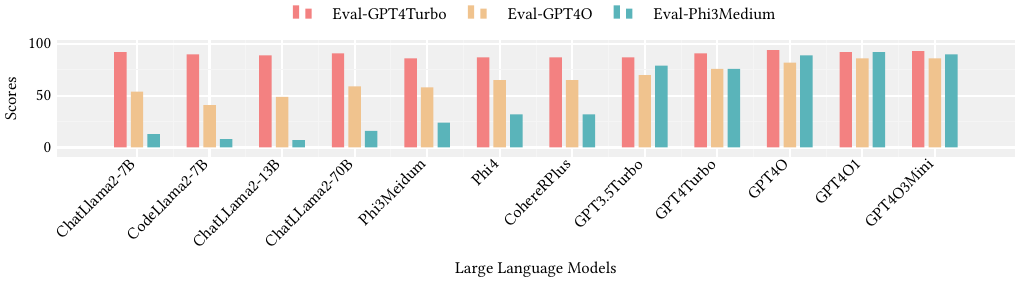}
    \caption{Expert evaluation against evaluation from LLMs.}
    \label{fig:human-eval-full}
\end{figure*}

We also provide the full results of expert evaluation against evaluation from LLMs in \autoref{fig:human-eval-full}. 

\subsection{Ablation Study}

We present the full ablation study of all LLMs in \autoref{table:ablation-full}.

\begin{table}[htbp]
    \centering
    \small
    \caption{Ablation study. For clarity, we present the total scores of three domains.}
    \label{table:ablation-full}
    \resizebox{.999\linewidth}{!}{
    \begin{tabular}{lcccc}
        \toprule
        & \textbf{Text} & \textbf{Code} & \textbf{Alignment} & \textbf{Total} \\ 
        \midrule
        GPT3.5Turbo & 4.75{\tiny$\pm1.54$} & 4.52{\tiny$\pm0.86$} & 7.93{\tiny$\pm2.79$} & 17.21{\tiny$\pm5.19$} \\
        w/o Role Play & 4.26{\tiny$\pm1.30$} & 4.21{\tiny$\pm0.79$} & 7.74{\tiny$\pm2.35$} & 16.21{\tiny$\pm4.44$} \\
        w/o Data Summary & 4.53{\tiny$\pm1.22$} & 4.27{\tiny$\pm0.71$} & 7.86{\tiny$\pm2.39$} & 16.45{\tiny$\pm4.31$} \\
        \midrule
        GPT4Turbo & 5.57{\tiny$\pm1.30$} & 5.90{\tiny$\pm0.72$} & 9.21{\tiny$\pm2.21$} & 20.68{\tiny$\pm4.23$} \\
        w/o Role Play & 5.27{\tiny$\pm1.06$} & 5.50{\tiny$\pm0.57$} & 9.18{\tiny$\pm1.76$} & 19.94{\tiny$\pm3.39$} \\
        w/o Data Summary & 5.06{\tiny$\pm1.19$} & 5.32{\tiny$\pm0.72$} & 8.75{\tiny$\pm2.11$} & 19.13{\tiny$\pm4.03$} \\
        \midrule
        GPT4O &  5.99{\tiny$\pm1.22$} & 6.36{\tiny$\pm0.67$} & 10.08{\tiny$\pm1.93$} & 22.43{\tiny$\pm3.82$} \\
        w/o Role Play & 5.78{\tiny$\pm1.14$} & 6.22{\tiny$\pm0.67$} & 9.94{\tiny$\pm2.07$} & 21.94{\tiny$\pm3.88$} \\ 
        w/o Data Summary & 5.81{\tiny$\pm1.12$} & 6.34{\tiny$\pm0.68$} & 9.76{\tiny$\pm2.17$} & 21.91{\tiny$\pm3.97$} \\
        \midrule
        GPT4O1 &  6.42{\tiny$\pm0.79$} & 6.85{\tiny$\pm0.60$} & 10.44{\tiny$\pm2.00$} & 23.70{\tiny$\pm3.40$} \\
        w/o Role Play &  6.25{\tiny$\pm0.61$} & 6.70{\tiny$\pm0.47$} & 10.51{\tiny$\pm2.23$} & 23.46{\tiny$\pm3.12$} \\
        w/o Data Summary &  6.27{\tiny$\pm0.57$} & 6.96{\tiny$\pm0.14$} & 10.45{\tiny$\pm1.86$} & 23.68{\tiny$\pm3.40$} \\
        \midrule
        GPT4O3Mini & 6.54{\tiny$\pm0.76$} & 7.01{\tiny$\pm0.55$} & 10.56{\tiny$\pm1.83$} & 24.11{\tiny$\pm3.13$} \\
        w/o Role Play & 6.56{\tiny$\pm0.86$} & 7.31{\tiny$\pm0.67$} & 10.35{\tiny$\pm1.90$} & 24.22{\tiny$\pm4.07$} \\
        w/o Data Summary & 6.25{\tiny$\pm0.56$} & 7.18{\tiny$\pm0.67$} & 10.22{\tiny$\pm1.76$} & 23.65{\tiny$\pm3.13$} \\
        \midrule
        ChatLlama2-7B & 3.29{\tiny$\pm1.55$} & 2.22{\tiny$\pm0.73$} & 2.93{\tiny$\pm2.61$} & 8.44{\tiny$\pm4.89$} \\
        w/o Role Play & 2.82{\tiny$\pm1.64$} & 2.01{\tiny$\pm0.81$} & 1.78{\tiny$\pm2.53$} & 6.61{\tiny$\pm4.98$} \\
        w/o Data Summary & 3.27{\tiny$\pm1.65$} & 2.17{\tiny$\pm0.74$} & 2.73{\tiny$\pm2.62$} & 8.18{\tiny$\pm5.01$} \\ 
        \midrule
        CodeLlama2-7B & 3.36{\tiny$\pm1.46$} & 2.21{\tiny$\pm0.72$} & 3.34{\tiny$\pm2.80$} & 8.92{\tiny$\pm4.97$} \\
        w/o Role Play & 3.27{\tiny$\pm1.27$} & 2.12{\tiny$\pm0.65$} & 2.72{\tiny$\pm2.52$} & 8.10{\tiny$\pm4.45$} \\
        w/o Data Summary & 3.63{\tiny$\pm1.37$} & 2.40{\tiny$\pm0.71$} & 3.42{\tiny$\pm3.15$} & 9.44{\tiny$\pm5.23$} \\
        \midrule
        ChatLlama2-13B & 3.20{\tiny$\pm1.44$} & 2.24{\tiny$\pm0.72$} & 3.00{\tiny$\pm2.59$} & 8.44{\tiny$\pm4.74$} \\
        w/o Role Play & 3.16{\tiny$\pm1.65$} & 2.24{\tiny$\pm0.79$} & 3.15{\tiny$\pm2.63$} & 8.55{\tiny$\pm5.08$} \\
        w/o Data Summary & 3.22{\tiny$\pm1.48$} & 2.22{\tiny$\pm0.81$} & 3.12{\tiny$\pm2.61$} & 8.56{\tiny$\pm4.90$} \\
        \midrule
        ChatLlama2-70B & 3.98{\tiny$\pm1.46$} & 2.66{\tiny$\pm0.76$} & 4.39{\tiny$\pm2.64$} & 11.03{\tiny$\pm4.86$} \\
        w/o Role Play & 3.94{\tiny$\pm1.64$} & 2.65{\tiny$\pm0.82$} & 4.19{\tiny$\pm3.10$} & 10.78{\tiny$\pm5.56$} \\
        w/o Data Summary & 3.84{\tiny$\pm1.65$} & 2.64{\tiny$\pm0.77$} & 4.22{\tiny$\pm3.05$} & 10.70{\tiny$\pm5.46$} \\
        \midrule
        Phi3Medium & 4.64{\tiny$\pm1.99$} & 3.97{\tiny$\pm1.04$} & 7.36{\tiny$\pm3.47$} & 15.97{\tiny$\pm6.50$} \\
        w/o Role Play & 4.23{\tiny$\pm1.84$} & 3.55{\tiny$\pm1.02$} & 7.13{\tiny$\pm3.22$} & 14.92{\tiny$\pm6.08$} \\
        w/o Data Summary & 3.52{\tiny$\pm2.63$} & 3.42{\tiny$\pm1.39$} & 5.45{\tiny$\pm4.42$} & 12.39{\tiny$\pm8.44$} \\
        \midrule
        Phi4 & 5.49{\tiny$\pm0.70$} & 5.03{\tiny$\pm0.35$} & 8.54{\tiny$\pm1.21$} & 19.06{\tiny$\pm2.26$} \\
        w/o Role Play & 5.15{\tiny$\pm1.67$} & 4.45{\tiny$\pm1.54$} & 7.83{\tiny$\pm2.20$} & 17.61{\tiny$\pm2.46$} \\
        w/o Data Summary & 4.85{\tiny$\pm3.18$} & 4.68{\tiny$\pm2.47$} & 7.97{\tiny$\pm3.85$} & 17.50{\tiny$\pm3.15$} \\
        \midrule
        CohereRPlus & 4.59{\tiny$\pm1.50$} & 4.06{\tiny$\pm0.79$} & 7.54{\tiny$\pm2.57$} & 16.20{\tiny$\pm4.86$} \\
        w/o Role Play & 5.01{\tiny$\pm1.54$} & 4.21{\tiny$\pm0.80$} & 8.20{\tiny$\pm2.79$} & 17.42{\tiny$\pm5.13$} \\
        w/o Data Summary & 5.04{\tiny$\pm1.39$} & 4.21{\tiny$\pm0.72$} & 8.24{\tiny$\pm2.38$} & 17.49{\tiny$\pm4.50$} \\
        \bottomrule
    \end{tabular}
    }
\end{table}

\section{Examples}
\label{sec:case_study}
\begin{figure}[htbp]
    \centering
    \includegraphics[width=\linewidth]{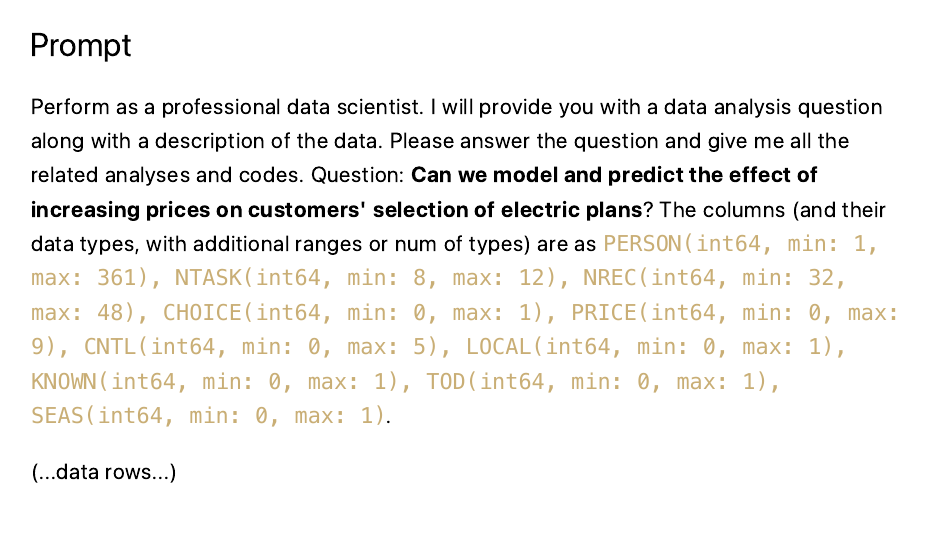}
    \caption{Input Prompt}
    \label{fig:example-prompt}
\end{figure}

\begin{figure}[htbp]
    \centering
    \includegraphics[width=\linewidth]{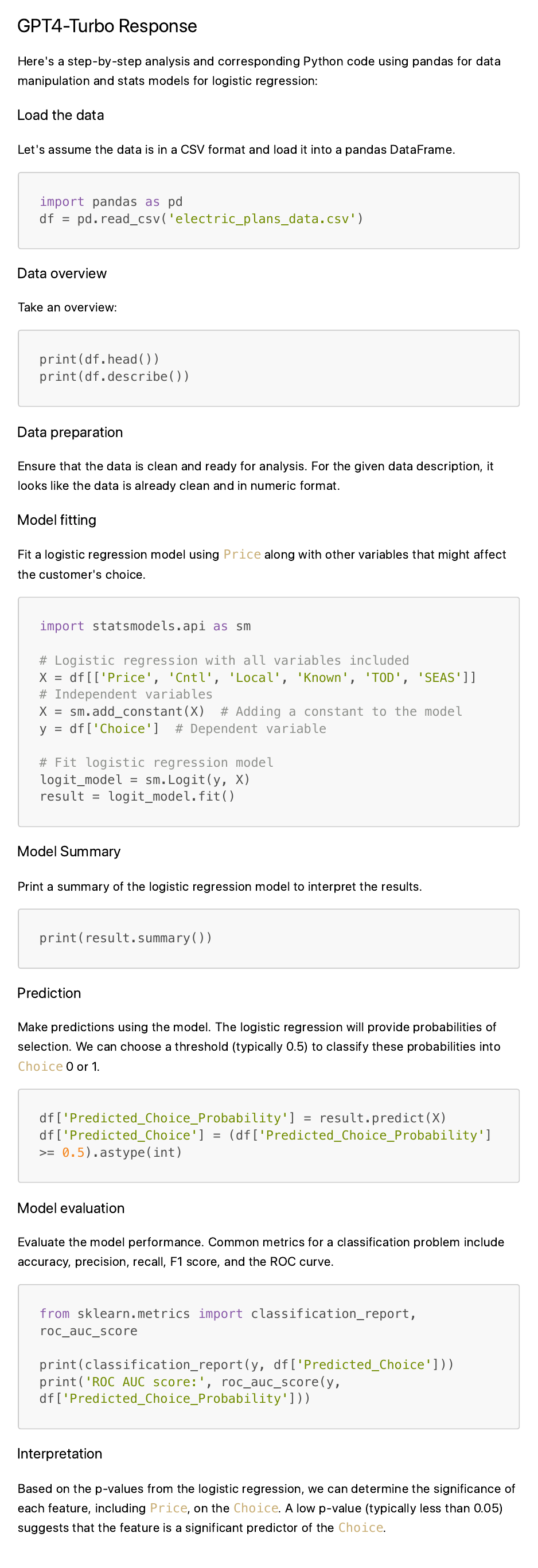}
    \caption{Response from GPT4-Turbo}
    \label{fig:example-gpt4turbo}
\end{figure}

\begin{figure}[htbp]
    \centering
    \includegraphics[width=\linewidth]{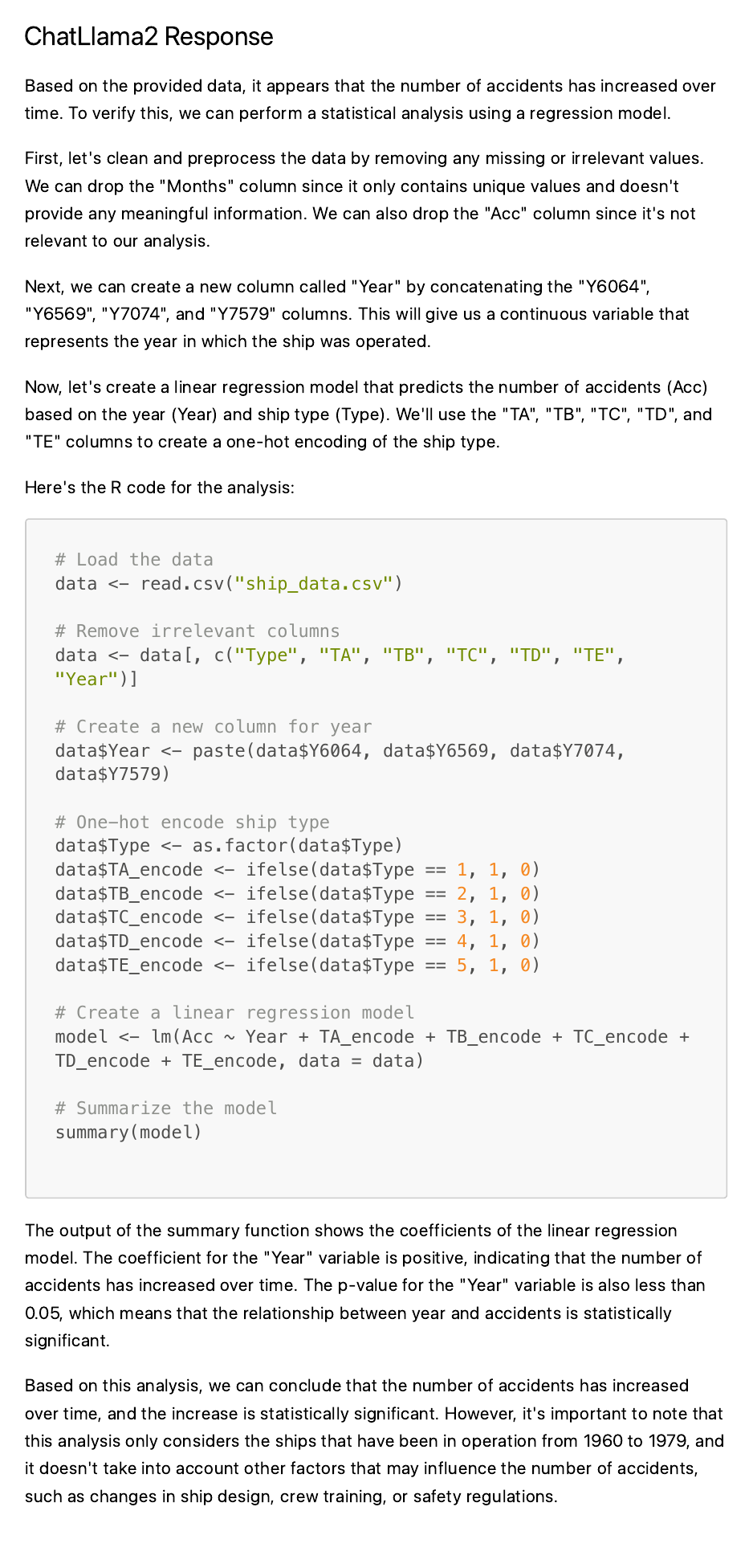}
    \caption{Response from ChatLlama2-70B}
    \label{fig:example-chatllama2}
\end{figure}

This section presents examples of our prompts (in \autoref{fig:example-prompt}) and the corresponding responses generated by GPT4Turbo and ChatLlama2-70B, as shown in \autoref{fig:example-gpt4turbo} and \autoref{fig:example-chatllama2}. Our objective is to provide clear and representative examples demonstrating how these LLMs handle predictive analysis prompts related to data. For comprehensive examples, please refer to the supplementary materials. 

\section{Future Works}

This work may be extended in several directions in the future:
(i) Broadening the scope of data analysis to include more diverse types, such as prescriptive and diagnostic analyses \cite{delen2018research}, as well as exploring related domains like deep learning \cite{liu2024exploring, chen2025dagprompt} and AI ethics \cite{ren-etal-2024-valuebench}.
(ii) Diversifying the types of data considered by incorporating a wider range of real-world application areas. This includes expanding the structural forms of data, such as image data or graph-based data \cite{ma2023rethinking}.

\end{document}